%% file: new_arxiv_version_paper.tex
\documentclass[10pt,logo,onecolumn,copyright]{nvidiatechreport}

\usepackage{mdframed}
\usepackage[utf8]{inputenc} 
\usepackage[T1]{fontenc}    

\usepackage{amsfonts}       
\usepackage{nicefrac}       
\usepackage{microtype}      
\usepackage[dvipsnames]{xcolor}         
\usepackage[dvipsnames]{xcolor}         
\definecolor{mediumgray}{gray}{0.6}
\usepackage{multirow}
\usepackage{multirow}
\usepackage{multicol}
\usepackage{graphicx}
\usepackage[numbers]{natbib}
\usepackage{tabto}
\usepackage{xspace}
\usepackage{amsmath}
\usepackage{adjustbox}
\usepackage{enumitem}
\usepackage{wrapfig}
\usepackage{dblfloatfix}

\usepackage{footmisc}

\usepackage{float}

\usepackage{natbib}

\usepackage{hyperref}
\usepackage{url}
\usepackage{booktabs}
\usepackage{multirow}
\usepackage{enumitem}
\usepackage{adjustbox}
\usepackage{tabularx}
\usepackage{arydshln}
\usepackage{wrapfig}

\usepackage{multirow}
\usepackage{colortbl}
\usepackage{amsmath}
\usepackage{makecell}
\usepackage{hhline}
\usepackage{array}
\usepackage{diagbox}
\usepackage{graphicx}
\usepackage{adjustbox}

\usepackage{fp} 

\usepackage{xcolor}
\usepackage[subtle]{savetrees}

\usepackage{subfig}
\usepackage{booktabs}
\usepackage{multirow}
\usepackage{cleveref}

\usepackage{pifont}
\definecolor{baselinecolor}{gray}{.9}

\usepackage{color, colortbl}
\definecolor{Gray}{gray}{0.9}

\makeatletter
\DeclareRobustCommand\onedot{\futurelet\@let@token\@onedot}
\def\@onedot{\ifx\@let@token.\else.\null\fi\xspace}

\def\etal{\emph{et al}\onedot}
\makeatother

\title{SDUM: A Scalable Deep Unrolled Model for Universal MRI Reconstruction}

\author{Puyang Wang\textsuperscript{1,$*$} \;
Pengfei Guo\textsuperscript{2,$*$} \; 
Keyi Chai\textsuperscript{1} \; 
Jinyuan Zhou\textsuperscript{1} \;
Daguang Xu\textsuperscript{2} \;
Shanshan Jiang\textsuperscript{1} \\~\\
\textsuperscript{1}Johns Hopkins University \quad \textsuperscript{2}NVIDIA    \\
\textsuperscript{$*$}Equal contribution}

\begin{abstract}
Clinical cardiac MRI spans diverse contrasts, sampling trajectories, accelerations, scanners, and patient populations, yet most deep-learning reconstruction methods remain protocol specific.
We present the Scalable Deep Unrolled Model (SDUM), which integrates a Restormer-based unrolled reconstructor, per-cascade coil sensitivity estimation, sampling aware weighted data consistency, and universal conditioning on cascade index and acquisition metadata.
A single SDUM model achieves state-of-the-art performance across all CMRxRecon2025 tracks without task-specific fine-tuning and outperforms PromptMR+ on CMRxRecon2024 by ${+}0.55$~dB.
Scaling experiments show near-logarithmic gains with depth up to 18 cascades ($r{=}0.986$, $R^2{=}0.973$) and continued but diminishing gains from data scaling (32.72~dB at 40\% to 33.18~dB at 100\%).
SDUM also generalizes in a zero-shot setting to unseen in-house CEST MRI (43.57~dB PSNR, 0.9769 SSIM).
When trained separately on fastMRI brain, SDUM surpasses PC-RNN by ${+}1.8$~dB.
These results support SDUM as a scalable framework for robust MRI reconstruction across heterogeneous acquisition settings beyond cardiac MRI.

\smallskip
\textbf{Code:}
\href{https://github.com/NVIDIA-Medtech/NV-Raw2insights-MRI}{https://github.com/NVIDIA-Medtech/NV-Raw2insights-MRI}

\textbf{Model:}
\href{https://huggingface.co/nvidia/NV-Raw2insights-MRI}{https://huggingface.co/nvidia/NV-Raw2insights-MRI}

\end{abstract}

\begin{document}



\maketitle



\input{sec/1_intro}
\input{sec/2_related_work}
\input{sec/3_method}

\input{sec/4_results}
\input{sec/5_conclusion}

\noindent\textbf{Acknowledgments.} This study was supported partially by grants from the NIH (R01AG06917, R01CA276221, and R37CA248077).

\bibliographystyle{splncs04}
\bibliography{main_eccv}
\input{sec/X_suppl}



\end{document}

%% file: sec/1_intro.tex
\section{Introduction}
\label{sec:intro}

Cardiac MRI (CMR) is the gold standard for non-invasive assessment of cardiac morphology, function, and tissue characterization. A single clinical CMR exam routinely acquires a dozen or more sequences, cine for wall motion, T1/T2 mapping for tissue characterization, late gadolinium enhancement (LGE) for fibrosis, perfusion for ischemia, flow for valvular assessment, tagging for strain, using Cartesian, radial, spiral, or kt-space sampling at acceleration factors ranging from 4$\times$ to 24$\times$. Scanner field strengths (1.5T, 3T, emerging 5T), coil arrays, vendor software, and patient populations (adult, pediatric, multi-disease) add further heterogeneity across clinical sites. While recent deep learning methods~\cite{hammernik2018learning,schlemper2017deep,duan2019vs,aggarwal2019modl,sriram2020end} have delivered strong performance in controlled settings, they typically fracture along protocol boundaries: a model trained for one sampling mask or acceleration often degrades when the acquisition changes. This brittleness hinders deployment in multi-site clinical environments and motivates a \emph{single, universally applicable} cardiac MRI reconstruction model that can robustly handle heterogeneous inputs without per-protocol retraining.

\textbf{Hypothesis.} We hypothesize that despite the exceptional diversity of cardiac MRI protocols, the underlying cardiac anatomy and physiology impose a \emph{shared image prior} that can be learned by a single model. Specifically, a deep unrolled reconstruction network, when conditioned on the acquisition physics (sampling pattern, trajectory type, acceleration factor) via universal conditioning and trained with a progressive curriculum, can disentangle protocol-specific acquisition artifacts from anatomy-specific content. The shared prior across protocols acts as implicit regularization, enabling robustness to distribution shifts across centers, diseases, field strengths, and patient populations.

\textbf{The scaling gap.} In adjacent domains, empirical scaling analyses~\cite{kaplan2020scaling,hoffmann2022training,zhai2022scaling}, quantifying how error decays with parameters, data, or compute,have become a practical compass for model design and resource allocation. MRI reconstruction lacks such guidance. Practitioners must rely on trial-and-error to decide whether to invest in wider backbones, deeper cascade unrolling, or larger training sets, with no principled understanding of returns or saturation.

\textbf{Limitations of existing approaches.} Current attempts at universal MRI reconstruction remain incomplete. Many deep unrolled models~\cite{hammernik2018learning,aggarwal2019modl} employ a scalar data-consistency (DC) weight that ignores sampling-density and noise characteristics specific to each trajectory. Conditioning mechanisms, when present, are often ad hoc or task-specific~\cite{xin2024enhanced}. Coil sensitivity maps (CSMs) are typically precomputed and fixed, limiting robustness to motion and field inhomogeneities. No existing approach has been validated as a single model across the full spectrum of cardiac MRI heterogeneity: multi-contrast, multi-trajectory, multi-center, multi-field-strength, and multi-population, simultaneously.

\textbf{Our approach: SDUM.} We introduce the Scalable Deep Unrolled Model (SDUM), a framework designed for universal cardiac MRI reconstruction through five synergistic components:
\begin{enumerate}
\item \textbf{Restormer-based reconstruction backbone}: Multi-Dconv Head Transposed Attention (MDTA) and Gated-Dconv Feedforward Network (GDFN)~\cite{zamir2022restormer} jointly capture long-range dependencies (to unfold aliasing) and local structures (to preserve edges) with favorable memory efficiency. A shallow two-stage pyramid preserves high-resolution detail while enabling global context aggregation.

\item \textbf{Learned per-cascade coil sensitivity map estimation (CSME)}: A U-Net-based estimator refines CSMs at each cascade, mitigating errors from motion, noise, and field inhomogeneities without relying on autocalibration regions~\cite{pruessmann1999sense,uecker2014espirit}.

\item \textbf{Sampling-aware weighted data consistency (SWDC)}: Instead of a scalar DC weight, we learn spatially varying k-space weight maps conditioned on the sampling pattern, enabling mask-specific and density-aware fidelity enforcement that unifies Cartesian and non-Cartesian trajectories in a single learnable module.

\item \textbf{Universal conditioning (UC)}: Sinusoidal embeddings of cascade index $t$ and protocol metadata (mask type, acceleration, modality) are mapped through MLPs and injected into every Restormer block~\cite{perez2018film,peebles2023scalable}, allowing a single model to adapt its behavior across diverse acquisitions.

\item \textbf{Progressive cascade expansion}: Training proceeds by doubling interior cascades while fixing endpoints, stabilizing optimization and reusing learned weights as depth increases~\cite{bengio2009curriculum,karras2018progressive}.
\end{enumerate}

We train SDUM on heterogeneous CMRxRecon data covering cardiac protocols (BlackBlood, Cine, Flow2d, LGE, Mapping, Perfusion, T1rho, T1w, T2w, Aorta, Tagging) with Cartesian and non-Cartesian sampling. On the CMRxRecon2025 challenge~\cite{Xu_2025_CMRxRecon2025}, which evaluates generalization to unseen centers, diseases, 5T field strength, and pediatric populations, a \emph{single} SDUM achieves state-of-the-art performance across all four tracks without task-specific fine-tuning, exceeding specialized baselines by up to ${+}1.0$~dB. On CMRxRecon2024~\cite{wang2025towards}, SDUM surpasses the winning method PromptMR+~\cite{xin2024enhanced} by ${+}0.55$~dB and wins in over 90\% of paired validation cases. We further evaluate strict zero-shot transfer on in-house CEST MRI from an unseen scanner vendor/protocol, where SDUM reaches 43.57~dB PSNR and 0.9769 SSIM without adaptation. To examine whether SDUM's design principles transfer beyond cardiac MRI, we train a separate model on fastMRI brain data~\cite{zbontar2018fastmri}, where it exceeds PC-RNN~\cite{chen2022pyramid} by ${+}1.8$~dB. We also conduct (to our knowledge) the first \textbf{empirical scaling analysis for CMR reconstruction}, sweeping cascade depth $T$ from 1 to 18 and training data volume from 40\% to 100\%. Reconstruction quality follows PSNR~${\sim}$~$\log$(parameters) with correlation $r{=}0.986$. Beyond reporting numbers, we distill practical scaling lessons for building MRI reconstruction foundation models, including the need for diversity-aware data scaling and compute-optimal training.

%% file: sec/2_related_work.tex
\section{Related Work}
\label{sec:related}

\textbf{Classical and deep learning MRI reconstruction.}
Classical accelerated MRI reconstruction employs compressed sensing~\cite{lustig2008compressed}, parallel imaging (SENSE~\cite{pruessmann1999sense}, GRAPPA~\cite{griswold2002grappa}, ESPIRiT~\cite{uecker2014espirit}), and dynamic imaging methods~\cite{feng2014golden,otazo2015low}. Deep unrolled models (DUMs)~\cite{schlemper2017deep,hammernik2018learning,duan2019vs,aggarwal2019modl,sriram2020end} integrate iterative optimization into neural networks, alternating data-consistency (DC) and learned regularization steps. Related approaches include KIKI-Net~\cite{eo2018kiki}, DeepComplexMRI~\cite{wang2020deepcomplexmri}, PC-RNN~\cite{chen2022pyramid}, and Plug-and-Play/RED frameworks~\cite{venkatakrishnan2013plug,romano2017little,zhang2017beyond}. Most unrolled models use scalar DC weights; our SWDC learns spatially varying k-space weights conditioned on sampling patterns. E2E-VarNet~\cite{sriram2020end} jointly learns reconstruction and sensitivity estimation, inspiring our learned CSME approach.

\noindent\textbf{Cardiac MRI reconstruction.}
Cardiac MRI poses unique challenges due to the diversity of contrasts (cine, mapping, LGE, perfusion, flow), dynamic acquisitions requiring temporal modeling, and the need for robustness across field strengths and patient populations. Recent cardiac-focused methods include PromptMR~\cite{xin2023fill} and PromptMR+~\cite{xin2024enhanced}, which use prompt-based conditioning for multi-contrast cardiac reconstruction, and HierAdaptMR~\cite{xu2025hieradaptmr}, which employs hierarchical adaptation across centers and protocols. The CMRxRecon challenge series~\cite{wang2025towards,Xu_2025_CMRxRecon2025} has catalyzed progress by providing large-scale multi-contrast, multi-trajectory cardiac k-space datasets and standardized evaluation across clinically relevant distribution shifts. Despite this progress, no prior method has demonstrated a single model that generalizes across the full spectrum of cardiac MRI heterogeneity---multi-contrast, multi-trajectory, multi-center, multi-field-strength, and multi-population---without task-specific fine-tuning.

\noindent\textbf{Architectures and conditioning.}
U-Net~\cite{ronneberger2015u} and restoration transformers (SwinIR~\cite{liang2021swinir}, Restormer~\cite{zamir2022restormer}) are widely used backbones; we adopt Restormer for its favorable memory/compute trade-off. Conditional normalization (FiLM~\cite{perez2018film}, AdaLN~\cite{peebles2023scalable}) enables task adaptation; in MRI, PromptMR+~\cite{xin2024enhanced} and HierAdaptMR~\cite{xu2025hieradaptmr} use conditioning for multi-contrast reconstruction. We extend this paradigm with physics-informed universal conditioning.

\noindent\textbf{Scaling behavior and training.}
Empirical scaling analyses~\cite{kaplan2020scaling,hoffmann2022training,zhai2022scaling} quantify performance vs.\ parameters/data/compute in NLP and vision, guiding resource allocation. Prior work has shown that increasing unrolled depth (number of cascades) is important for reconstruction performance: ReconFormer~\cite{guo2023reconformer} demonstrated this with recurrent transformers, and similar findings were reported for deep unrolled models~\cite{xin2024rethinking}. However, no prior study has systematically characterized the scaling behavior of unrolled MRI models across both depth and data axes. For training, our cascade expansion applies progressive training~\cite{bengio2009curriculum,karras2018progressive} to stabilize optimization during deep unrolling.

%% file: sec/3_method.tex
\section{Methods}

\subsection{Problem Formulation}
MRI reconstruction aims to recover an image $x \in \mathbb{C}^{H \times W}$ from undersampled multi-coil $k$-space measurements $y \in \mathbb{C}^{C \times H \times W}$ with $C$ receiver coils. The forward model is
\begin{equation}
    y = M \cdot \mathcal{F}(S \cdot x) + n,
\end{equation}
where $S$ denotes coil sensitivity maps, $\mathcal{F}$ is the Fourier transform, $M$ is a binary sampling mask, and $n$ is acquisition noise. Real-world cardiac MRI protocols vary in contrast (cine, mapping, LGE, perfusion, flow), sampling pattern (Cartesian, radial, spiral, kt), and acceleration, so the model must handle heterogeneous acquisition conditions.

Reconstruction is posed as $\hat{x} = \arg\min_{x} \bigl( \frac{1}{2} \| M \mathcal{F}(S x) - y \|_2^2 + \lambda R(x) \bigr)$, balancing a data-consistency (DC) term with a regularizer $R(x)$.

\paragraph{Deep Unrolled Reconstruction.}
Let $A = M \cdot \mathcal{F}$. Starting from the zero-filled image $x^{(0)}$, cascade $t \in \{0,\dots,T-1\}$ performs proximal-gradient unrolling:
\begin{align}
z^{(t)} &= x^{(t)} - \tau_t S^{H} A^{H}\big(AS \cdot x^{(t)} - y\big) , \label{eq:pgd-grad}\\
x^{(t+1)} &= D_{\theta}^{(t)}\big(z^{(t)}\big), \label{eq:pgd-prox}
\end{align}
where $\tau_t$ is a learnable step size and $D_{\theta^{(t)}}$ is a learnable proximal operator. The full model $\hat{x} = x^{(T)}$ is trained end-to-end with conditioning on cascade index and protocol metadata.

\subsection{Scalable Deep Unrolled Models}

\begin{figure}[t]
    \centering
    \includegraphics[width=\linewidth]{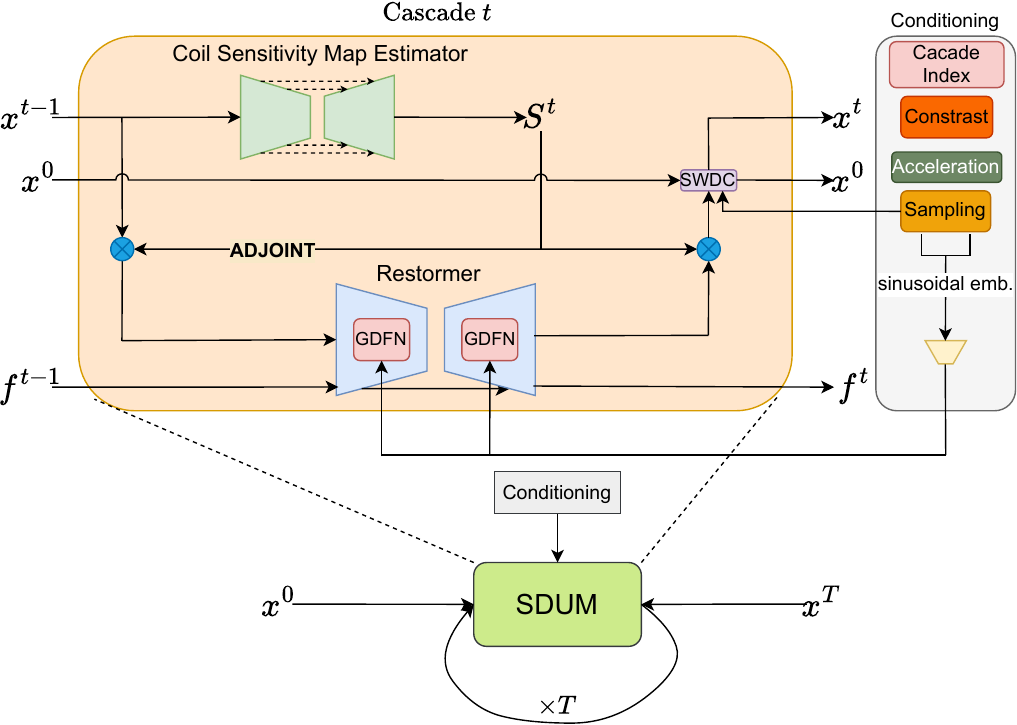}
  \caption{\textbf{SDUM overview.} Each cascade combines a Restormer-based reconstructor, a learned CSME, sampling-aware weighted data consistency (SWDC), and universal conditioning (UC) on cascade index and protocol metadata.}
\label{fig:overview}
\end{figure}

Deep unrolled MRI reconstruction methods often rely on precomputed coil sensitivity maps (CSMs), which are inevitably imperfect due to noise, motion, and field inhomogeneities.

We adopt a deep-unrolled architecture with $T$ cascades. Each cascade comprises (i) a Restormer-based image reconstructor, (ii) a U-Net-based coil-sensitivity map estimator (CSME) that refines CSMs during reconstruction and does not rely on ACS regions, and (iii) a sampling-aware data-consistency operator that enforces the forward model with mask-dependent weighting in k-space. We introduce two key enhancements:

\textbf{Adaptive unrolling with cascade skip connections.} Each cascade receives not only the output of the previous stage, but also skip connections of decoder feature at downsampling level $f_l$ from previous cascade. This mitigates vanishing gradients, facilitates training deeper models, and supports effective reuse of intermediate features.

\textbf{Adjacent slice/frame input.} For dynamic (eg. cinematic cardiac MRI) and 3D volumetric MRI, each cascade can access temporally or spatially adjacent frames. These inputs are fed into the reconstructor alongside the current slice, enabling the model to leverage temporal redundancy or 3D anatomical context.

\subsection{Restormer Backbone}
\label{sec:backbone}
We adopt Restormer~\cite{zamir2022restormer} as the per-cascade reconstructor because its Multi-Dconv Head Transposed Attention (MDTA) and Gated-Dconv Feedforward Network (GDFN) jointly capture non-local and local dependencies at $O(HW\!\cdot\!C^2)$ complexity. We use a shallow two-stage pyramid (one down/up) with MDTA+GDFN blocks, which preserves high-resolution detail while providing global context. Empirically, this shallow design outperforms deeper pyramids in our setting: additional downsampling increases receptive field but tends to oversmooth fine structures and weaken small-lesion contrast. A wider-but-shallow per-cascade design also stabilizes deep unrolling by improving gradient flow across cascades. Additional implementation details are provided in the supplementary material.

At cascade $t\!\in\!\{0,\dots,T\!-\!1\}$, we estimate coil sensitivities with a U-Net CSME and then apply DC and a Restormer prior:
\begin{align}
S^{(t)} &= \mathrm{Norm}\!\left(\mathrm{CSME}_{\phi_t}\big(x^{(t)}\big)\right), \label{eq:csme}\\
z^{(t)} &= x^{(t)} - \tau_t\, {S^{(t)}}^{\!H} A^{H}\!\left(A\!\big(S^{(t)} \cdot x^{(t)}\big) - y\right), \label{eq:pgd-grad-rest}\\
x^{(t+1)} &= \mathrm{Restormer}_{\theta_t}\!\big(z^{(t)}\big), \label{eq:pgd-prox-rest}
\end{align}
where $A{=}\,M\mathcal{F}$, $\tau_t$ is a learnable DC step size, and $\mathrm{Norm}(\cdot)$ enforces per-pixel coil normalization ($\sum_c |S_c|^2{=}1$).

Each cascade has its own CSME and Restormer parameters.
The CSME is a compact complex U-Net with channel progression $\{12,24,48,96,192\}$,
instance normalization, and deconvolution-based upsampling; its output is normalized by
$S \leftarrow S / \sqrt{\sum_c |S_c|^2}$ to keep physically meaningful coil-energy scaling.
For the image prior, we use a two-level Restormer with channel widths $\{256,512\}$,
attention heads $\{1,2\}$, and block depths $\{3,6\}$, followed by two refinement blocks.
This shallow-but-wide design preserves high-resolution structure while still enabling long-range
interaction through MDTA. Complex-valued MRI data are represented as stacked real/imaginary
channels and processed with real-valued operators, and inputs are padded to sizes divisible by the
down/up-sampling hierarchy for numerically stable unrolling.

\subsection{Sampling-aware Weighted Data Consistency}

We replace the standard scalar DC weight with a learned, spatially varying k-space weight map conditioned on the sampling pattern. The standard DC gradient in cascade $t$ is
\begin{equation}
    g^{(t)}_{\mathrm{DC}} \;=\; {S^{(t)}}^{\!H} A^{H}\!\left( A\!\big(S^{(t)} \!\cdot\! x^{(t)}\big) - y \right),
    \label{eq:dc_grad_standard}
\end{equation}
leading to the update $z^{(t)} = x^{(t)} - \tau_t\, g^{(t)}_{\mathrm{DC}}$.

\begin{figure}[t]
    \centering
    \includegraphics[width=\linewidth]{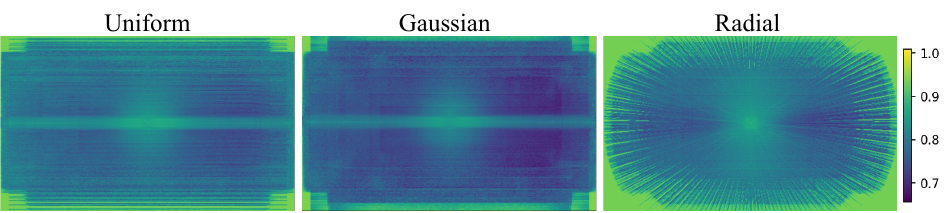}
  \caption{\textbf{Example learned sampling-aware data-consistency} weight maps showing spatial adaptation to sampling patterns.}
\label{fig:swdc}
\end{figure}

\paragraph{Weighted residual in $k$-space.} Let $w^{(t)} \in \mathbb{R}^{H\times W}$ be a nonnegative weight map defined in $k$-space (broadcast across coils), with $w^{(t)}\!=\!0$ on unsampled locations. We form a weighted residual
\begin{equation}
    r^{(t)}_{\!w} \;=\; w^{(t)} \odot \!\left( A\!\big(S^{(t)} \!\cdot\! x^{(t)}\big) - y \right),
\end{equation}
and replace \eqref{eq:dc_grad_standard} by
\begin{align}
    g^{(t)}_{\mathrm{SWDC}} \;&=\; {S^{(t)}}^{\!H} A^{H}\!\big( r^{(t)}_{\!w} \big)
    \label{eq:swdc_grad}
\end{align}
yielding the SWDC update $z^{(t)} \;=\; x^{(t)} - \tau_t\, g^{(t)}_{\mathrm{SWDC}}$.
When $w^{(t)}\!=\!M$, \eqref{eq:swdc_grad} recovers the standard DC. As shown in \cref{fig:swdc}, the SWDC module learns distinct weight maps tailored to uniform, Gaussian, and radial sampling patterns. The learnable weight $w^{(t)}$ is initialized at maximum resolution $H{\times}W$ and center-cropped to match $y$; we maintain distinct $w_s^{(t)}$ for each sampling pattern $s$ to enable pattern-specific DC. A detailed justification of SWDC over classical density compensation is provided in the supplementary material.

\subsection{Universal Conditioning Mechanism}

\label{sec:universal_conditioning}
We condition every cascade on (i) its cascade index $t$ and (ii) a discrete protocol label $y$ encoding mask type, acquisition type, and acceleration factor. Both are mapped by sinusoidal embeddings followed by two-layer MLPs and \emph{summed} to form a single conditioning vector injected into all transformer blocks, enabling a single model to adapt across cascade depth and protocol.

Let $\phi(\cdot)\!:\!\mathbb{R}\!\to\!\mathbb{R}^{d_0}$ denote a sinusoidal encoder and let $h_t,h_y:\mathbb{R}^{d_0}\!\to\!\mathbb{R}^{d}$ be two-layer MLPs,
\begin{align}
c   &= h_t\!\big(\phi(t)\big) + h_y\!\big(\phi(y)\big) \in \mathbb{R}^{d}.
\end{align}

\textbf{Injection into Transformer blocks.} Let $f^{(t)}_{\ell}\!\in\!\mathbb{R}^{H_\ell\times W_\ell\times C_\ell}$ be features at layer $\ell$ in cascade $t$. Each transformer block receives $c$ and applies attention and a gated depthwise FFN (GDFN) with an \emph{additive, spatially broadcast} bias derived from $c$:
\begin{align}
\tilde f^{(t)}_{\ell} &= \mathrm{MDTA}\!\big(f^{(t)}_{\ell}\big) \;+\; f^{(t)}_{\ell},\\
f^{(t)}_{\ell+1}      &= \mathrm{GDFN}\!\big(\tilde f^{(t)}_{\ell}\big)\;+\;B_\ell(c),
\end{align}
where $B_\ell(c)$ produces a $C_\ell$-dimensional vector that is broadcast across $(H_\ell,W_\ell)$ using single linear layer. The same $c$ is passed (via a timestep-aware sequential wrapper) to \emph{all} encoder, decoder, and refinement blocks of each cascade in the Restormer.

\subsection{Progressive Cascade Expansion}
We train SDUM with a curriculum that progressively grows the cascade depth~\cite{bengio2009curriculum,karras2018progressive}. From a stage with $T_{k-1}$ cascades, the next stage expands to $T_k = 2\,(T_{k-1}-1)$ by keeping the first and last cascades fixed and duplicating only the interior cascades. All parameters (Restormer weights, CSME weights, DC step sizes, SWDC weights) are warm-started from the previous stage. This ``end-fixed, middle-doubling'' schedule refines the interior discretization, where residual correction is largest, without disturbing already-converged endpoint parameters, improving gradient flow and convergence at large $T$.

Concretely, we define a stage-transition index map
\[
\pi_k(t)=
\begin{cases}
0, & t=0,\\
T_{k-1}-1, & t=T_k-1,\\
1+\lfloor (t-1)/2 \rfloor, & \text{otherwise,}
\end{cases}
\]
so each new cascade $t\in\{0,\dots,T_k-1\}$ copies parameters from cascade $\pi_k(t)$ at the previous stage.
In practice, this means only interior cascades are duplicated, while endpoint cascades are inherited unchanged:
\begin{align}
\theta^{(t)} &\leftarrow \theta_{\mathrm{prev}}^{(\pi_k(t))}, &
\phi^{(t)} &\leftarrow \phi_{\mathrm{prev}}^{(\pi_k(t))}, \nonumber\\
\tau_t &\leftarrow \tau_{\mathrm{prev}}^{(\pi_k(t))}, &
w^{(t)} &\leftarrow w_{\mathrm{prev}}^{(\pi_k(t))}.
\end{align}
Starting from a base depth $T_0{=}6$, the schedule used in this work yields $T{=}6\rightarrow10\rightarrow18$,
which empirically provides stable optimization and consistent gains over direct one-shot training at $T{=}18$.

%% file: sec/4_results.tex
\section{Experiments}
\label{sec:experiments}
We evaluate SDUM on \textbf{CMRxRecon2024}~\cite{wang2025towards}, \textbf{CMRxRecon2025}~\cite{Xu_2025_CMRxRecon2025}, and \textbf{fastMRI multi-coil brain}~\cite{zbontar2018fastmri}. CMRxRecon2025 extends CMRxRecon2024 by evaluating reconstruction performance under distribution shifts arising from unseen centers, disease categories, field strengths, and pediatric cohorts. Unless stated otherwise, results are reported on official validation splits or challenge leaderboards and measured by SSIM~$\uparrow$, PSNR~$\uparrow$, and NMSE~$\downarrow$.

\subsection{Baselines}
On fastMRI brain, we compare against classical CS~\cite{lustig2008compressed}, U-Net~\cite{ronneberger2015u}, D5C5~\cite{schlemper2017deep}, KIKI-Net~\cite{eo2018kiki}, VN~\cite{hammernik2018learning}, VS-Net~\cite{duan2019vs}, ComplexMRI~\cite{wang2020deepcomplexmri}, and PC-RNN~\cite{chen2022pyramid} (fastMRI winner). For CMRxRecon2024/2025, we follow official protocols and compare against top leaderboard methods, including PromptMR+~\cite{xin2024enhanced}, Hamedani \etal~\cite{anvari2024all}, vSHARP+ARN~\cite{yiasemis2024deep}, IMR~\cite{wang2025towards}, Shen \etal~\cite{shen_synapse_profile_2025}, HierAdaptMR~\cite{xu2025hieradaptmr}, and GENRE-CMR~\cite{hamedani2025genre}.

\input{tables/cmrxrecon25_results}
\input{tables/cmrxrecon24_results}

\subsection{Comparisons with State-of-the-Art}\label{sec:sota}

\noindent\textbf{CMRxRecon2025: Universal cardiac MRI reconstruction.} \cref{tab:cmrxrecon25} reports results on the CMRxRecon2025 leaderboard, which evaluates model generalization across unseen centers, diseases, field strengths, and age groups. A \emph{\textbf{single SDUM model}} ($T{=}18$) is used for all four subtasks without any task-specific fine-tuning, and it achieves the highest SSIM and PSNR throughout, surpassing strong baselines including PromptMR+ (\textit{32 cas.})~\cite{xin2024enhanced}, HierAdaptMR~\cite{xu2025hieradaptmr}, Shen \etal~\cite{shen_synapse_profile_2025}, and GENRE-CMR~\cite{hamedani2025genre}. 
On Regular Task~1 (\textbf{multi-center generalization}), where both validation and test data come from entirely unseen medical centers, SDUM attains an absolute gain of +0.26~dB over the second-best method. On Regular Task~2 (\textbf{multi-disease evaluation}), which assesses robustness to unseen cardiovascular conditions with undisclosed disease labels, SDUM maintains superior performance. For the two special tracks (\textbf{5T} and \textbf{Pediatric}), SDUM generalizes to field strengths and age distributions absent from training, exceeding the next-best methods by up to +1.0~dB PSNR. These results validate our hypothesis: a single model conditioned on protocol physics can generalize across the full spectrum of cardiac MRI heterogeneity.

\begin{figure}[t]
    \centering
    \includegraphics[width=\linewidth]{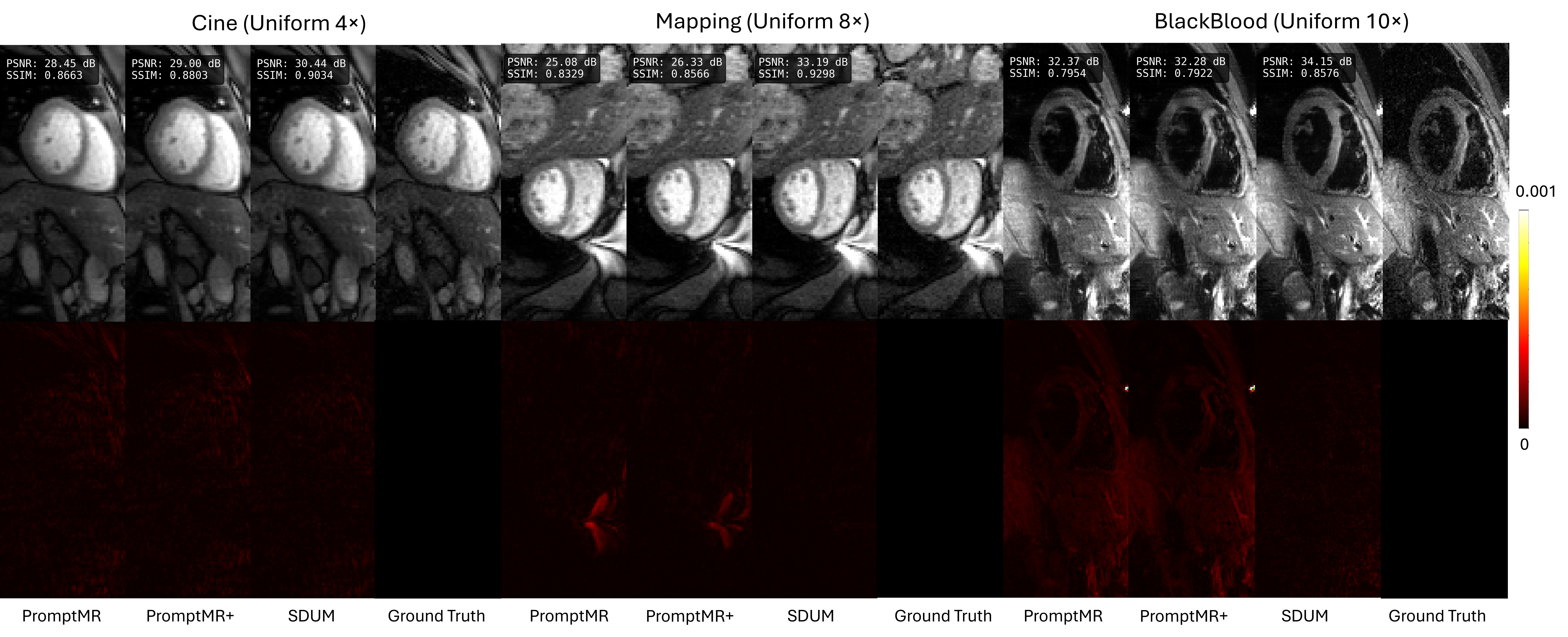}
  \caption{\textbf{Qualitative comparison across cardiac imaging modalities.}
  Three representative cases from CMRxRecon2024 Task~1 validation set show the superior performance of SDUM over PromptMR and PromptMR+.
  }
\label{fig:qualitative}
\end{figure}

\noindent\textbf{CMRxRecon2024.} \cref{tab:cmrxrecon24} reports results on the official CMRxRecon2024 leaderboard across two subtasks. SDUM outperforms the 2024 winning method PromptMR+~\cite{xin2024enhanced} by 0.55~dB in Task~1 and provides consistently higher fidelity under both spatially uniform and temporally varying sampling conditions. A paired-case analysis over the full validation set further shows that SDUM beats PromptMR+ in 90.3\% of Task~1 cases and 93.9\% of Task~2 cases, with mean gains of +1.091~dB and +1.060~dB, respectively. Qualitative examples (\cref{fig:qualitative}) show clearer myocardial boundaries and fewer residual aliasing artifacts.

\subsection{Ablation Study}\label{sec:ablation}

We conduct ablations on the CMRxRecon2025 challenge validation leaderboard. Results are summarized in \cref{tab:ablation}.

\input{tables/ablation_combine}

\noindent\textbf{Backbone variants.} We evaluate different reconstruction backbones, U-Net~\cite{ronneberger2015u}, DiT~\cite{peebles2023scalable}, TAU~\cite{tan2023temporal}, and Restormer~\cite{zamir2022restormer} using $T{=}6$ cascades with approximately the same number of trainable parameters. Restormer achieves the best overall performance (backbone block in \cref{tab:ablation}(a)) by coupling efficient global attention with strong local modeling.

\noindent\textbf{Down-sampling depth.} We vary the number of down-sampling layers within each SDUM cascade with $T{=}6$. As shown in the DS block of \cref{tab:ablation}(b), two layers provide the best balance between contextual coverage and spatial fidelity (32.09~dB). Fewer layers hinder abstraction (training fails at one layer), while deeper hierarchies lose fine texture fidelity.

\noindent\textbf{CSME variants.} We compare using a single shared CSME versus per-cascade estimators. The CSME block in \cref{tab:ablation}(d) shows that the multi-CSME configuration improves PSNR by ${+}0.51$~dB, confirming the value of progressive sensitivity refinement.

\noindent\textbf{Data consistency formulations.} We ablate different DC layers: simple learnable DC, weighted DC (WDC), and our SWDC. The DC block in \cref{tab:ablation}(e) shows that SWDC achieves the best result, outperforming learnable DC by ${+}0.43$~dB.

\noindent\textbf{UC and iterative behavior.} The UC/iteration block in \cref{tab:ablation}(f) shows that enabling UC consistently improves quality (${+}0.38$~dB). Three cascades without iterative refinement outperform a single cascade with three iterations at similar compute, indicating that unrolled depth is more beneficial than additional iterations under limited compute.

\subsection{Scaling Analysis}\label{sec:scaling}

\begin{figure}[t]
    \centering
    \includegraphics[width=0.6\linewidth]{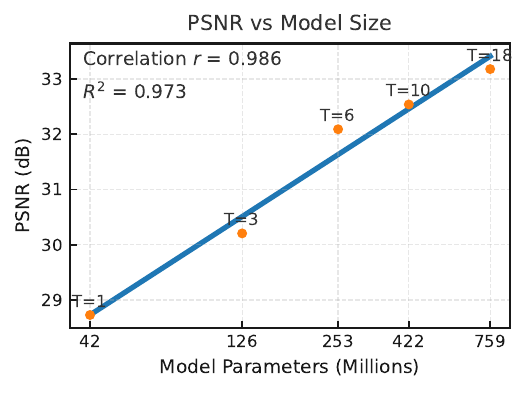}
  \caption{\textbf{Scaling behavior of unrolled depth $T$.} Reconstruction PSNR follows an approximately linear relationship with $\log$(\# parameters), with $r{=}0.986$ and $R^2{=}0.973$ up to $T{=}18$.}
\label{fig:scaling-params}
\end{figure}

\noindent\textbf{Width scaling.} We adjust network width from 7.2M to 78.7M parameters while fixing $T{=}1$. The single-cascade capacity block in \cref{tab:ablation}(c) shows that performance improves steadily up to ${\sim}42$M parameters but saturates thereafter, suggesting that depth $T$ yields more efficient capacity scaling than width.

\noindent\textbf{Depth scaling.} \cref{fig:scaling-params} shows the scaling trend with respect to unrolled depth~$T$.
As the number of cascades increases from 1 to 18 (42\,M to 759\,M parameters), PSNR improves monotonically, following an approximately linear relationship with $\log(\text{\# parameters})$ ($r{=}0.986$, $R^2{=}0.973$).
This predictable scaling provides practical guidance for trading compute against reconstruction fidelity.

\noindent\textbf{Data scaling.} \cref{tab:data_scaling} reports SDUM ($T{=}18$) trained on 40\%, 80\%, and 100\% of the training data and evaluated on CMRxRecon2025 Regular Task~1. Performance improves consistently from 32.72 to 33.18~dB PSNR, from 0.882 to 0.895 SSIM, and from 0.016 to 0.014 NMSE.

\input{tables/data_scaling}

The gains are monotonic but not uniform: 40\%$\rightarrow$80\% yields +0.33~dB, while 80\%$\rightarrow$100\% yields +0.13~dB. This indicates diminishing marginal returns with scale, yet no saturation at the current regime. For the community, this suggests that simply adding more samples helps, but scaling should increasingly emphasize \emph{data diversity} (vendors, trajectories, pathologies, field strengths, and age groups) rather than only increasing volume within already-dominant distributions.

\noindent\textbf{Inference efficiency.} On a single NVIDIA H100 GPU, the full $T{=}18$ model completes inference in ${\sim}1.0$~s using ${\sim}6$~GB memory at $256{\times}256$, and under 4~s with ${\sim}11$~GB peak memory at $328{\times}768$, making it practical for near--real-time clinical deployment (detailed hardware settings are in the supplementary material).

\subsection{Cross-Anatomy Generalization: fastMRI Brain}\label{sec:fastmri}

To verify that SDUM's design principles transfer beyond cardiac MRI, we train a \emph{separate} model on fastMRI brain data~\cite{zbontar2018fastmri} with $T{=}6$ cascades. \cref{tab:fastmri_brain} reports results on the fastMRI multi-coil brain benchmark at $4\times$ and $6\times$ acceleration. 
SDUM consistently outperforms prior methods, exceeding the strong recurrent baseline PC-RNN~\cite{chen2022pyramid} by +1.8~dB PSNR, validating the scalable unrolled Restormer backbone and SWDC.
These results demonstrate that the architectural design, while developed and validated primarily for cardiac MRI, generalizes to a different anatomy and dataset, suggesting the approach is not overfit to cardiac-specific characteristics.

\input{tables/fastmri_result}

\subsection{Zero-Shot Reconstruction of CEST MRI}
\label{sec:cest}

\begin{figure}[t]
    \centering
    \includegraphics[width=\linewidth]{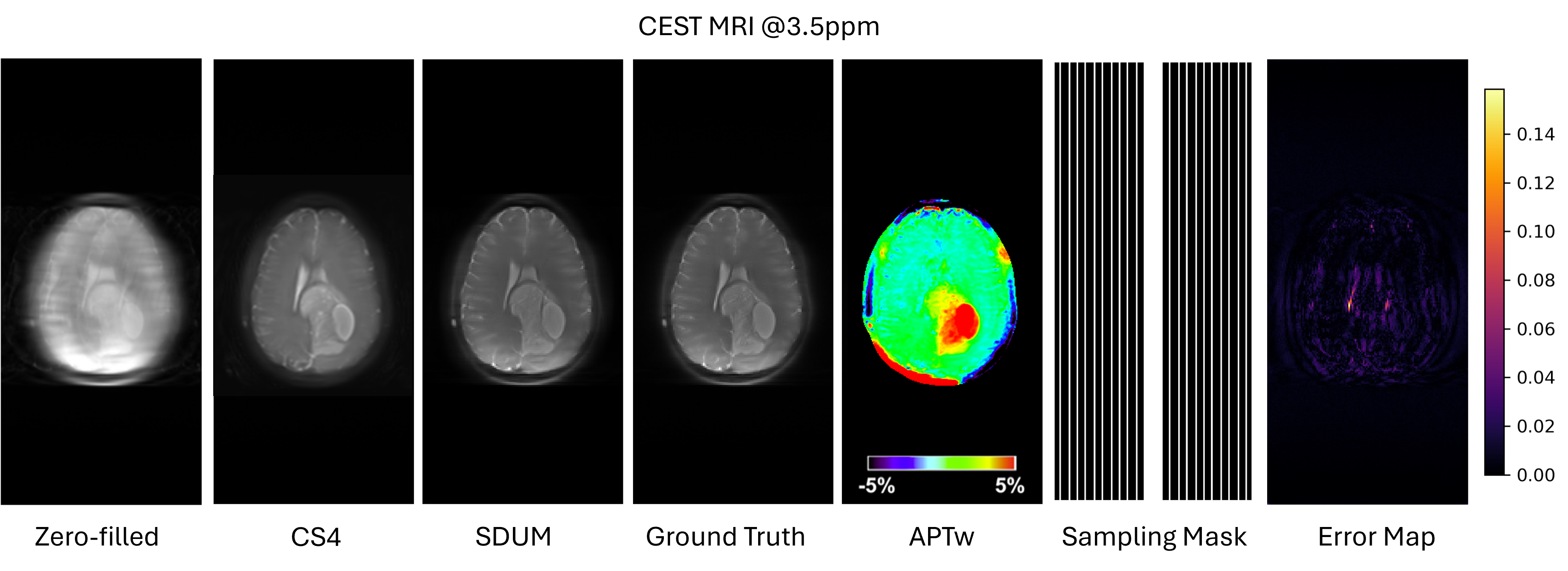}
  \caption{\textbf{Zero-shot CEST MRI reconstruction on in-house data.}
  Left to right: zero-filled, scanner CS4, SDUM, fully sampled reference, APTw map ($\pm$\,3.5\,ppm), sampling mask, and absolute error. SDUM reaches 43.57\,dB PSNR and 0.9769 SSIM without fine-tuning.
  }
\label{fig:cest}
\end{figure}

For out-of-distribution evaluation, we apply SDUM trained on fastMRI brain directly to in-house CEST MRI from brain-tumor patients acquired on a 3T Philips scanner, with retrospectively $4\times$ undersampled Cartesian multi-coil $k$-space. Because this vendor, pulse sequence, and image contrast are absent from training, this is a strict zero-shot test. As shown in \cref{fig:cest}, SDUM suppresses strong aliasing and recovers finer structures and lesion contrast more faithfully than zero-filled and scanner CS4 reconstructions. Quantitatively, SDUM reaches 43.57\,dB PSNR and 0.9769 SSIM against the fully sampled reference while preserving downstream APTw-map fidelity, indicating robust cross-vendor and cross-protocol generalization.

%% file: tables/cmrxrecon25_results.tex
\begin{table}[!t]
\centering
\caption{\textbf{Quantitative comparison on the CMRxRecon2025 challenge.} Results are reported on the official leaderboard across four subtasks. A single SDUM ($T{=}18$) is used for all tracks without task-specific fine-tuning.}
\label{tab:cmrxrecon25}
\scriptsize
\setlength{\tabcolsep}{2.5pt}
\renewcommand{\arraystretch}{1.03}
\begin{tabular}{lccc|lccc}
\hline
\multicolumn{4}{c|}{\textbf{Regular Task 1: Multi-center}} & \multicolumn{4}{c}{\textbf{Regular Task 2: Multiple diseases}} \\
Method & SSIM & PSNR & NMSE & Method & SSIM & PSNR & NMSE \\ \hline
HierAdaptMR~\cite{xu2025hieradaptmr} & 0.870 & 31.702 & 0.019 & HierAdaptMR~\cite{xu2025hieradaptmr} & 0.864 & 32.520 & 0.021 \\
Shen \etal~\cite{shen_synapse_profile_2025} & 0.878 & 32.19 & 0.017 & Shen \etal~\cite{shen_synapse_profile_2025} & 0.872 & 32.998 & 0.017 \\
PromptMR+~\cite{xin2024enhanced} & 0.891 & 32.919 & 0.014 & PromptMR+~\cite{xin2024enhanced} & 0.879 & 33.422 & 0.014 \\
\textbf{SDUM (Ours)} & \textbf{0.895} & \textbf{33.179} & \textbf{0.014} & \textbf{SDUM (Ours)} & \textbf{0.880} & \textbf{33.538} & \textbf{0.014} \\ \hline
\multicolumn{4}{c|}{\textbf{Special Task 1: 5T}} & \multicolumn{4}{c}{\textbf{Special Task 2: Pediatric imaging}} \\
Method & SSIM & PSNR & NMSE & Method & SSIM & PSNR & NMSE \\ \hline
HierAdaptMR~\cite{xu2025hieradaptmr} & 0.888 & 33.290 & 0.018 & GENRE-CMR~\cite{hamedani2025genre} & 0.872 & 31.656 & 0.033 \\
Shen \etal~\cite{shen_synapse_profile_2025} & 0.893 & 33.605 & 0.017 & HierAdaptMR~\cite{xu2025hieradaptmr} & 0.880 & 32.014 & 0.030 \\
PromptMR+~\cite{xin2024enhanced} & 0.895 & 33.824 & 0.016 & Shen \etal~\cite{shen_synapse_profile_2025} & 0.887 & 32.436 & 0.028 \\
\textbf{SDUM (Ours)} & \textbf{0.901} & \textbf{34.225} & \textbf{0.015} & \textbf{SDUM (Ours)} & \textbf{0.905} & \textbf{33.478} & \textbf{0.020} \\ \hline
\end{tabular}
\end{table}

%% file: tables/cmrxrecon24_results.tex
\begin{table}[!t]
\centering
\caption{\textbf{Quantitative comparison on the CMRxRecon2024 challenge.} Results are reported on the official validation leaderboard across Task~1 and Task~2. SDUM is configured to $T{=}18$.}
\label{tab:cmrxrecon24}
\scriptsize
\setlength{\tabcolsep}{2pt}
\renewcommand{\arraystretch}{1.0}
\begin{tabular}{lccc|lccc}
\hline
\multicolumn{4}{c|}{\textbf{Task 1: Uniform} ($4\times, 8\times, 10\times$)} & \multicolumn{4}{c}{\textbf{Task 2: k-t Gauss./Rad./Unif.} ($4\times$--$24\times$)} \\
Method & SSIM & PSNR & NMSE & Method & SSIM & PSNR & NMSE \\ \hline
vSHARP w/ ARN~\cite{yiasemis2024deep} & 0.910 & 34.449 & 0.022 & IMR~\cite{zhang_profile_3472363} & 0.885 & 32.479 & 0.034 \\
Hamedani \etal~\cite{anvari2024all} & 0.912 & 34.522 & 0.022 & vSHARP w/ ARN~\cite{yiasemis2024deep} & 0.888 & 32.712 & 0.033 \\
PromptMR~\cite{xin2023fill} & 0.907 & 34.053 & 0.023 & PromptMR~\cite{xin2023fill} & 0.886 & 32.492 & 0.033 \\
PromptMR+ (\textit{12 cas.})~\cite{xin2024rethinking} & 0.915 & 34.677 & 0.020 & PromptMR+ (\textit{12 cas.})~\cite{xin2024rethinking} & 0.896 & 33.078 & 0.030 \\
PromptMR+ (\textit{32 cas.})~\cite{xin2024enhanced} & 0.921 & 35.152 & 0.018 & PromptMR+ (\textit{32 cas.})~\cite{xin2024enhanced} & 0.907 & 33.812 & 0.026 \\
\textbf{SDUM (Ours)} & \textbf{0.931} & \textbf{35.700} & \textbf{0.016} & \textbf{SDUM (Ours)} & \textbf{0.911} & \textbf{33.948} & \textbf{0.025} \\ \hline
\end{tabular}
\end{table}

%% file: tables/ablation_combine.tex
\begin{table}[!t]
	\centering
	\caption{\textbf{Ablation study on CMRxRecon2025 challenge.} Results are reported on the official leaderboard of Regular Task~1 (multi-center evaluation). Default configurations are marked in \colorbox{baselinecolor}{gray}.}
	\label{tab:ablation}
	\footnotesize
	\setlength{\tabcolsep}{3.0pt}
	\renewcommand{\arraystretch}{1.0}
	\begin{tabular}{@{}l@{\hspace{6pt}}>{\raggedright\arraybackslash}p{0.38\linewidth}>{\centering\arraybackslash}p{0.12\linewidth}>{\centering\arraybackslash}p{0.12\linewidth}>{\centering\arraybackslash}p{0.12\linewidth}@{}}
		\toprule
		\textbf{Study} & \textbf{Setting} & \textbf{SSIM} & \textbf{PSNR} & \textbf{NMSE} \\
		\midrule
		\multicolumn{5}{@{}l}{\textbf{(a) Backbone variants} ($T{=}6$)} \\
		& U-Net~\cite{ronneberger2015u} & 0.811 & 28.734 & 0.032 \\
		& DiT~\cite{peebles2023scalable} & 0.825 & 29.542 & 0.028 \\
		& TAU~\cite{tan2023temporal} & 0.873 & 31.597 & 0.018 \\
		\rowcolor{Gray} & Restormer~\cite{zamir2022restormer} & \textbf{0.881} & \textbf{32.090} & \textbf{0.017} \\
		\midrule
		
		\multicolumn{5}{@{}l}{\textbf{(b) Down-sampling (DS) layers}} \\
		& 4 & 0.872 & 31.544 & 0.019 \\
		& 3 & 0.877 & 31.787 & 0.018 \\
		\rowcolor{Gray} & 2 & \textbf{0.881} & \textbf{32.090} & \textbf{0.017} \\
		& 1 & \multicolumn{3}{c}{\emph{Training failed}} \\
		\midrule
		
		\multicolumn{5}{@{}l}{\textbf{(c) Single-cascade capacity scaling} (different widths, $T{=}1$)} \\
		& 7.2\,M params. & 0.784 & 27.605 & 0.042 \\
		& 13.4\,M params. & 0.803 & 28.271 & 0.036 \\
		\rowcolor{Gray} & 42.2\,M params. & 0.814 & 28.728 & 0.033 \\
		& 78.7\,M params. & \textbf{0.821} & \textbf{28.978} & \textbf{0.031} \\
		\midrule
		
		\multicolumn{5}{@{}l}{\textbf{(d) CSME variants} (\emph{Multiple}: one CSME per cascade)} \\
		& Single & 0.840 & 29.694 & 0.026 \\
		\rowcolor{Gray} & Multiple & \textbf{0.845} & \textbf{30.205} & \textbf{0.025} \\
		\midrule
		
		\multicolumn{5}{@{}l}{\textbf{(e) Data consistency (DC) layer variants} (WDC: Weighted DC)} \\
		& Simple Learnable & 0.872 & 31.661 & 0.018 \\
		& WDC & 0.877 & 31.787 & 0.017 \\
		\rowcolor{Gray} & SWDC & \textbf{0.881} & \textbf{32.090} & \textbf{0.017} \\
		\midrule
		
		\multicolumn{5}{@{}l}{\textbf{(f) Effect of UC and iterative reconstruction}} \\
		& $T{=}1$, Iter.=3, w/o UC  & 0.835 & 29.693 & 0.027 \\
		& $T{=}1$, Iter.=3, w/ UC & 0.843 & 30.071 & 0.025 \\
		\rowcolor{Gray} & $T{=}3$, Iter.=1, w/ UC  & \textbf{0.845} & \textbf{30.205} & \textbf{0.025} \\
		\bottomrule
	\end{tabular}
	
\end{table}

%% file: tables/data_scaling.tex
\begin{table}[t]
\centering
\caption{\textbf{Data scaling behavior.} SDUM ($T{=}18$) trained on different fractions of the training data, evaluated on CMRxRecon2025 Regular Task~1.}
\begin{tabular}{cccc}
\hline
\textbf{Data Fraction} & \textbf{SSIM} & \textbf{PSNR (dB)} & \textbf{NMSE} \\ \hline
40\% & 0.882 & 32.72 & 0.016 \\
80\% & 0.890 & 33.05 & 0.015 \\
100\% & \textbf{0.895} & \textbf{33.18} & \textbf{0.014} \\ \hline
\end{tabular}
\label{tab:data_scaling}
\end{table}

%% file: tables/fastmri_result.tex
\begin{table}[t]
\centering
\caption{\textbf{Quantitative comparison on fastMRI multi-coil brain.} We follow the setting of PC-RNN~\cite{chen2022pyramid} at $4\times$ and $6\times$ acceleration.} 
\label{tab:fastmri_brain}
\footnotesize
\setlength{\tabcolsep}{3pt}
\renewcommand{\arraystretch}{1.05}
\begin{tabular}{lcccc}
\hline
\multirow{2}{*}{Method} & \multicolumn{2}{c}{$4\times$} & \multicolumn{2}{c}{$6\times$} \\
 & SSIM & PSNR & SSIM & PSNR \\ \hline
CS~\cite{lustig2008compressed} & 0.904 & 35.2 & 0.851 & 31.2 \\
U-Net~\cite{ronneberger2015u} & 0.944 & 36.3 & 0.925 & 34.3 \\
VN~\cite{hammernik2018learning} & 0.938 & 36.2 & 0.900 & 32.6 \\
VS-Net~\cite{duan2019vs} & 0.950 & 37.6 & 0.921 & 34.3 \\
KIKI-Net~\cite{eo2018kiki} & 0.956 & 37.9 & 0.935 & 35.3 \\
D5C5~\cite{schlemper2017deep} & 0.955 & 38.0 & 0.931 & 34.9 \\
ComplexMRI~\cite{wang2020deepcomplexmri} & 0.953 & 37.9 & 0.933 & 35.2 \\
PC-RNN~\cite{chen2022pyramid} & 0.971 & 40.8 & 0.962 & 38.9 \\
\textbf{SDUM (Ours)} & \textbf{0.979} & \textbf{42.6} & \textbf{0.973} & \textbf{40.8} \\ \hline
\end{tabular}
\end{table}

%% file: sec/5_conclusion.tex
\section{Discussion}

SDUM integrates a Restormer backbone, per-cascade CSME, sampling-aware weighted data consistency, universal conditioning, and progressive cascade expansion within a single cardiac MRI reconstruction framework. Across CMRxRecon 2024/2025 tracks, the model maintains strong performance without task-specific fine-tuning, outperforms PromptMR+ on more than 90\% of paired CMRxRecon2024 validation cases, and shows zero-shot transfer to unseen-protocol CEST MRI (43.57~dB). Results on fastMRI (\(+1.8\)~dB over PC-RNN) further suggest potential generalization beyond the primary training domain.

The scaling experiments indicate that depth and data contribute differently to performance. PSNR follows an approximately logarithmic trend with parameter count up to $T{=}18$ ($r{=}0.986$), while gains from additional data diminish at higher fractions (40\%$\to$80\%: +0.33~dB; 80\%$\to$100\%: +0.13~dB). These observations motivate future work on compute-efficient scaling, diversity-aware data selection, and standardized conditioning metadata, alongside broader evaluation protocols that include downstream clinical consistency, robustness, and uncertainty calibration.

Several limitations remain. SWDC weight maps are not yet resolution-adaptive at inference, and training deeper variants is computationally expensive. In addition, the present scaling analysis does not fully characterize compute-optimal trade-offs. Although preliminary cross-anatomy results are encouraging, establishing a single anatomy-agnostic MRI reconstruction model requires further validation.

\section{Conclusion}

In summary, SDUM provides a single conditional reconstruction framework that achieves strong performance across multiple cardiac MRI reconstruction settings without task-specific fine-tuning. The observed depth and data scaling behavior indicates that further progress is likely to come from jointly optimizing model capacity, data diversity, and training compute. These findings support continued investigation of unified, scalable reconstruction models as a practical path toward more general MRI reconstruction systems.

%% file: sec/X_suppl.tex
\clearpage
\setcounter{section}{0}
\renewcommand{\thesection}{\Alph{section}}
\renewcommand{\thetable}{\Alph{section}.\arabic{table}}
\renewcommand{\thefigure}{\Alph{section}.\arabic{figure}}

\begin{center}
\Large\textbf{Supplementary Material}
\end{center}

\section{Additional Architecture Details}
\setcounter{table}{0}
\setcounter{figure}{0}

\subsection{Restormer Backbone Design}

Undersampling artifacts fold information over large spatial extents, while coil sensitivity and tissue boundaries induce fine local structure. A reconstructor must therefore (i) propagate \emph{global} context to untangle aliasing and (ii) preserve \emph{local} edges and textures to avoid hallucination. Purely convolutional U-Nets require depth or aggressive downsampling for long-range interactions, and windowed Transformers restrict cross-window communication. We adopt Restormer~\cite{zamir2022restormer} because its Multi-Dconv Head Transposed Attention (MDTA) and Gated-Dconv Feedforward Network (GDFN) jointly capture non-local and local dependencies with favorable memory and compute cost. Specifically, MDTA operates over $C\times C$ rather than $(HW)\times(HW)$, yielding complexity $O(HW\!\cdot\!C^2)$ while retaining non-local context, and its convolutional bias matches the piecewise-smooth statistics of MR images with sharp boundaries.

\textbf{Receptive field \& downsampling.}
Empirically, two stages with a single down/upsample yield higher PSNR/SSIM. High-resolution processing limits abstraction loss, while MDTA still provides long-range interaction. Deeper pyramids widen the receptive field but increase over-smoothing and hallucination risk.

\textbf{Unrolling efficiency.}
A shallow but wider per-cascade backbone stabilizes deep unrolling. It enables more cascades or larger input size without gradient accumulation, improving DC--prior interplay and gradient quality.

\textbf{MRI-specific choices.}
We use two channels (real/imag), residual scaling around blocks, and layer-wise normalization to handle scanner/protocol intensity shifts; no positional encodings are used.

\textbf{Implementation.}
We use two stages (one down, one up) with MDTA+GDFN blocks, long and short skips, and width tuned to fit maximum GPU memory, allowing large input K-space data for stable unrolled training.

\subsection{Progressive Cascade Expansion}

The motivation for progressive expansion is that, by analogy with ODE solvers, the endpoints of the unrolled chain (initial DC correction and final image refinement) tend to stabilize first during training, while the interior cascades carry most of the residual correction. By fixing the endpoints and doubling only interior cascades, we refine the discretization where it matters most without disturbing already-converged parameters.

Let $\theta^{(t)}$ denote proximal (Restormer/UNet) weights, $\phi^{(t)}$ the CSME weights, $\tau_t$ the DC step sizes, and $w^{(t)}$ the SWDC weights at cascade index $t$. Define the index mapping $\pi_k:\{0,\dots,T_k\!-\!1\}\!\to\!\{0,\dots,T_{k-1}\!-\!1\}$ as
\[
\pi_k(t) \;=\;
\begin{cases}
0, & t=0,\\[2pt]
T_{k-1}-1, & t = T_k-1,\\[2pt]
1 + \big\lfloor (t-1)/2 \big\rfloor, & \text{otherwise.}
\end{cases}
\]
Initialization at stage $k$ is then
\begin{align}
\theta^{(t)} &\leftarrow \theta_{\mathrm{prev}}^{(\pi_k(t))}, &
\phi^{(t)} &\leftarrow \phi_{\mathrm{prev}}^{(\pi_k(t))}, \nonumber\\
\tau_{t} &\leftarrow \tau^{(\pi_k(t))}_{\mathrm{prev}}, &
w^{(t)} &\leftarrow w_{\mathrm{prev}}^{(\pi_k(t))}. \label{eq:expansion-copy}
\end{align}
Thus, $t=0$ and $t=T_k\!-\!1$ copy the previous endpoints unchanged, while each interior cascade $j\in\{1,\dots,T_{k-1}-2\}$ is duplicated into two consecutive positions. With $T_k = 2(T_{k-1}-1)$, one can write $T_k = 2^k\,(T_0-2)+2$ for a given base depth $T_0\ge 2$.

\subsection{SWDC: Justification over Classical Approaches}

Classical Cartesian reconstruction assumes i.i.d.\ noise after pre-whitening~\cite{pruessmann1999sense}, while non-Cartesian methods apply analytically computed density compensation functions (DCFs)~\cite{pipe1999sampling}. SWDC subsumes both in a single learnable module for three reasons: (1)~after multi-coil combination the effective noise is spatially non-stationary, making uniform weighting suboptimal even for Cartesian data; (2)~analytic DCFs assume known, fixed sampling densities, whereas real acquisitions exhibit deviations due to gradient imperfections and variable-density designs; (3)~SWDC is jointly optimized with the reconstruction loss end-to-end, allowing the weighting to adapt not only to sampling density but also to the interplay with the learned prior across cascades. In our ablation (Tab.~1 in the main paper), SWDC provides a consistent ${+}0.43$~dB gain over a simple learnable scalar DC weight and ${+}0.30$~dB over a non-sampling-aware weighted DC, validating the benefit of learned, pattern-specific weighting.

\section{Dataset Details}
\setcounter{table}{0}
\setcounter{figure}{0}

We train SDUM on a heterogeneous combination of CMRxRecon2024, CMRxRecon2025, and fastMRI multi-coil brain datasets, covering a broad range of anatomical targets, contrasts, and sampling patterns.

\subsection{CMRxRecon2024} The CMRxRecon2024~\cite{wang2025towards} benchmark comprises two complementary tasks designed to evaluate contrast- and sampling-universal CMR reconstruction. Task~1 focuses on multi-contrast generalization under uniform undersampling, requiring a single model to reconstruct a broad set of cardiac MRI sequences---including Cine, Aorta, Mapping, Tagging, and the unseen Flow2D and BlackBlood contrasts---acquired on a Siemens 3\,T scanner. Data span multiple anatomical views (SAX, LAX, LVOT, aorta) and the published dataset includes 330 healthy volunteers (200 training, 60 validation, 70 test), totalling over 0.2 million k-space slices across six modalities. The released undersampling masks use 4$\times$, 8$\times$, and 10$\times$ acceleration (ACS excluded). Task~2 evaluates robustness to diverse sampling schemes, using a single mandatory model to handle uniform, Gaussian, and pseudo-radial trajectories across accelerations from 4$\times$ to 24$\times$; the same cohort split applies, and all data are multi-coil raw k-space. Together, these tasks assess whether a model can generalize across contrasts, views, sampling patterns, and acceleration regimes.

\subsection{CMRxRecon2025} The CMRxRecon2025~\cite{Xu_2025_CMRxRecon2025} challenge advances universal cardiac MRI reconstruction by evaluating model robustness to real clinical distribution shifts across four subtasks. Two Regular tasks assess multi-center/multi-vendor generalization and multi-disease robustness, while two Special tracks test adaptation to unseen field strength (5T) and pediatric imaging. The full dataset comprises 600 volunteers with multi-parametric CMR imaging, divided into 200 training cases (from 5 centers, including both healthy volunteers and patients), 100 validation cases (from the same 5 derivation centers plus 1 new center), and 300 test cases (from the 5 derivation centers and more than 5 additional unseen centers). These cases span diverse contrasts (Cine, T1/T2 mapping, LGE, perfusion), sampling trajectories (Cartesian, Gaussian, radial), and field strengths (1.5T, 3T, 5T). By requiring a single model to perform consistently across all subtasks without task-specific fine-tuning, CMRxRecon2025 pushes CMR reconstruction toward foundation-model-level generalization across centers, vendors, populations, and acquisition conditions.

\subsection{fastMRI} The fastMRI Brain dataset~\cite{zbontar2018fastmri} consists of 6,970 fully sampled multi-coil brain MRI scans acquired on 1.5T and 3T clinical scanners, covering four common contrasts: axial T1-weighted, post-contrast T1, T2-weighted, and FLAIR. The dataset includes 3,001 volumes at 1.5T and 3,969 volumes at 3T, each providing raw multi-coil k-space in a vendor-neutral format. As one of the largest public multi-coil brain MRI datasets, fastMRI Brain offers a realistic and diverse testbed for assessing cross-contrast and cross-field-strength generalization.

\section{Implementation Details}
\setcounter{table}{0}
\setcounter{figure}{0}

\subsection{Network Architecture}
Each cascade of SDUM consists of three tightly integrated components: a deep Restormer-based reconstructor, a coil-sensitivity map estimator (CSME), and a sampling-aware weighted data-consistency (SWDC) operator. 
The CSME is implemented as a Complex UNet with channel progression 
$\{12,24,48,96,192\}$, instance normalization, LeakyReLU activations, and deconvolution-based 
upsampling. The model operates on the fully sampled ACS region and outputs complex CSMs, 
which are explicitly normalized per spatial location using 
$x \leftarrow x / \sqrt{\sum_c |x_c|^2}$ to enforce $\sum_c |S_c|^2 = 1$, following 
standard SENSE practice.

The SDUM reconstructor is a two-level Restormer backbone. 
Each cascade receives a multi-frame complex input tensor (reshaped to
$2T$ channels) and processes it through a shallow $3{\times}3$ stem 
(10$\rightarrow$256 channels), followed by two hierarchical Transformer levels with 
channel widths $\{256, 512\}$, attention heads $\{1, 2\}$, and block depths 
$\{3, 6\}$. All Transformer Blocks employ MDTA, 
GDFN feed-forward layers with expansion factor~3, and LayerNorm-based 
pre-normalization with Drop Path ($p{=}0.1$). Downsampling and upsampling 
are performed via PixelUnshuffle/PixelShuffle, and skip connections propagate 
encoder features to the decoder at matching resolutions. A refinement stage with 
two additional Transformer Blocks further enhances the reconstructed output, followed 
by a $3{\times}3$ projection to restore multi-frame complex channels.

The reconstructor supports both \emph{time conditioning} (cascade index) and 
\emph{label conditioning} (sampling pattern, acceleration factor, and protocol type). 
Sinusoidal embeddings are transformed via a two-layer MLP and injected into every 
Transformer Block through additive channel-wise biases, enabling a shared model to 
adapt to heterogeneous acquisition settings. During unrolling, each cascade also receives cross-cascade skip features, allowing deeper cascades to reuse encoder/decoder outputs from previous cascades. SWDC is applied after 
each cascade using mask-specific learnable k-space weight maps, cropped to match the acquired resolution. The model operates on coil-reduced 
images. All operations use divisible padding aligned with the Restormer's 2-level downsampling hierarchy. Complex-valued MRI inputs are processed as stacked real and imaginary channels, and all convolutions are real-valued. 
Across the full model, 18 cascades are instantiated, each containing an independent CSME module and a full Restormer block stack, with inter-cascade skip connections.

\subsection{Training Procedure}
We follow the progressive cascade-expansion schedule outlined in the main paper, gradually increasing model depth from $T{=}6$ to $T{=}10$ and finally $T{=}18$, with interior cascades warm-started at each stage. Training uses the Muon optimizer~\cite{jordan2024muon} with a base learning rate of $2.4{\times}10^{-4}$, cosine decay, weight decay of $1{\times}10^{-3}$, and an SSIM reconstruction loss. Each stage is optimized for roughly 20k steps. We employ mixed-precision training (BF16) and gradient checkpointing to reduce memory consumption, and the batch size is adaptively selected based on the input resolution and available GPU memory.

All data processing and augmentation are implemented based on MONAI~\cite{cardoso2022monai} and performed directly in \emph{k-space}.
From the raw complex k-space input, we apply randomized transformations
including axis-aligned flips, random k-space shifts, k-space phase shifts,
gamma-based complex contrast augmentation, and spatially varying k-space
resize/crop, and simulated removal of readout oversampling.
A randomized under-sampling mask is then applied using the unified operator, which samples from multiple mask families, each with
independently randomized center fractions and acceleration factors.
The masked k-space is inverted to generate aliased images, and the ground-truth target is obtained by inverse-FFT of the fully sampled k-space.

\section{Additional Tables}
\setcounter{table}{0}
\setcounter{figure}{0}

Data scaling behavior is moved to the main paper (\cref{tab:data_scaling}).

\input{tables/training_infra}
\input{tables/inference_details}

\section{Additional Experiment Results}
\setcounter{table}{0}
\setcounter{figure}{0}

\subsection{Statistical Analysis and Comprehensive Comparison}

\begin{figure}[ht]
	\centering
	\includegraphics[width=\linewidth]{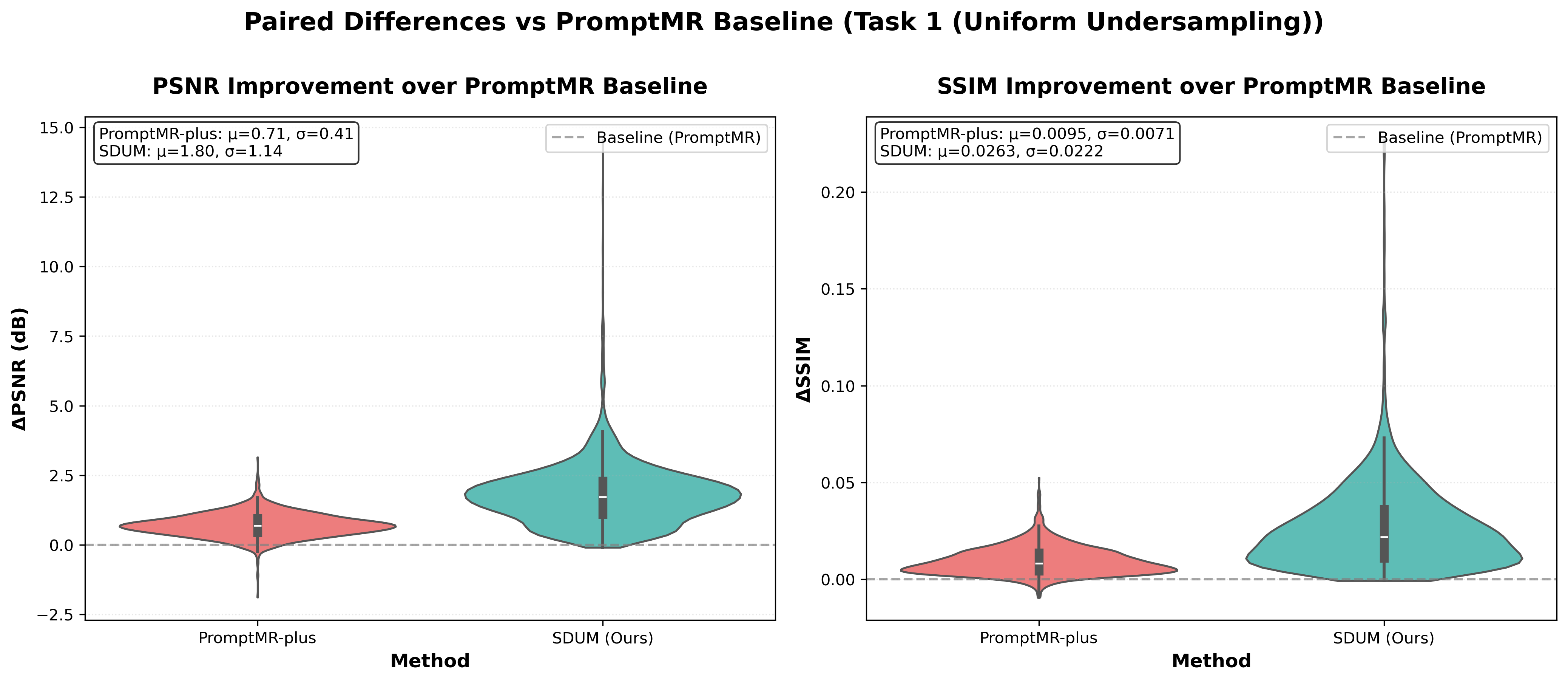}
	\caption{Statistical comparison of PSNR and SSIM improvements over PromptMR baseline for Task~1 (Uniform Undersampling).}
	\label{fig:violin_plot_task1}
\end{figure}

\begin{figure}[ht]
	\centering
	\includegraphics[width=\linewidth]{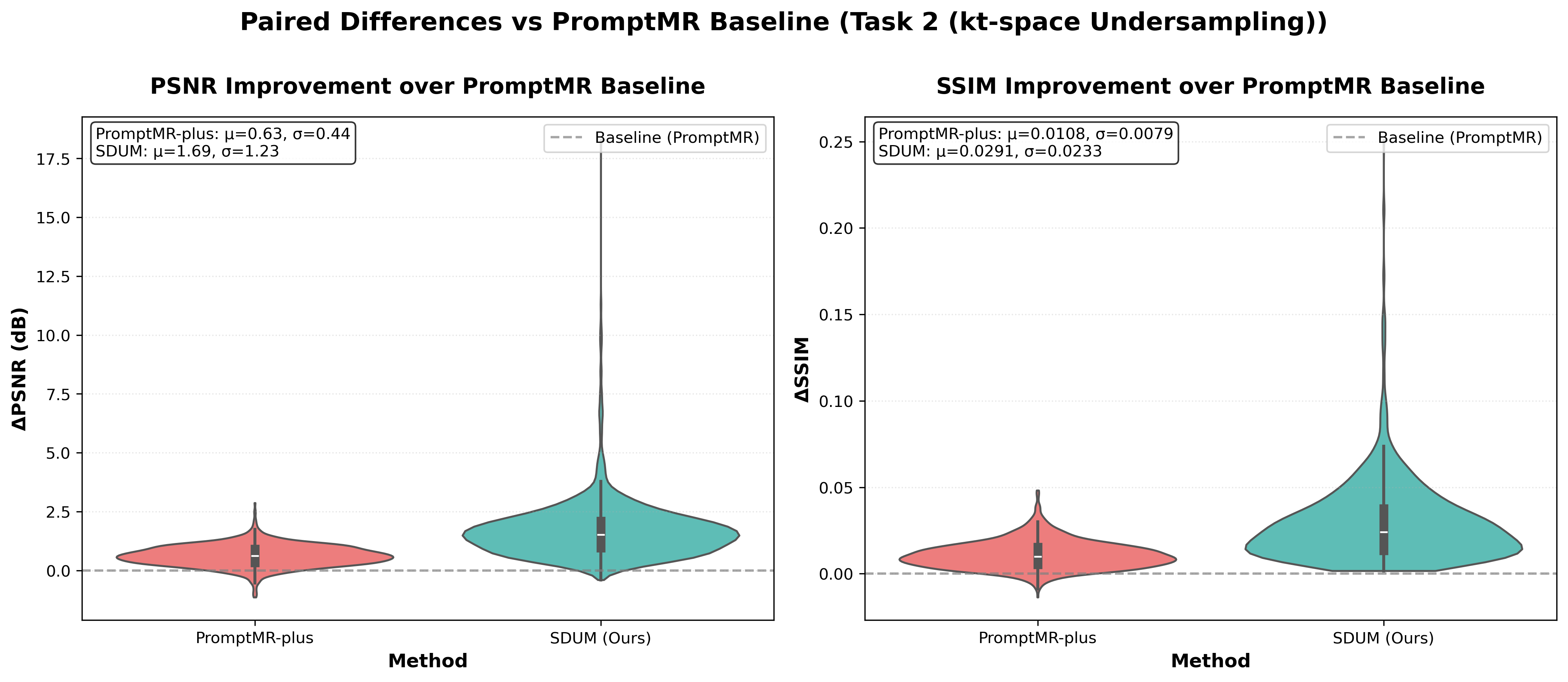}
	\caption{Statistical comparison of PSNR and SSIM improvements over PromptMR baseline for Task~2 (kt-space Undersampling).}
	\label{fig:violin_plot_task2}
\end{figure}

To provide a rigorous statistical assessment of SDUM, we conducted a comprehensive paired-difference analysis using PromptMR as the baseline across all validation cases from the CMRxRecon 2024 Challenge. \cref{fig:violin_plot_task1} and \cref{fig:violin_plot_task2} present violin plots illustrating the distributions of $\Delta$PSNR and $\Delta$SSIM improvements for both Task~1 (uniform undersampling, $n{=}1{,}614$) and Task~2 (kt-space undersampling, $n{=}1{,}281$). 

In Task~1, SDUM shows a mean PSNR improvement of $1.799 \pm 1.142$~dB versus $0.708 \pm 0.413$~dB for PromptMR+, representing a $1.091$~dB advantage. This superiority is even more pronounced in SSIM, where SDUM achieves $0.0263 \pm 0.0223$ improvement compared to $0.00946 \pm 0.00715$ for PromptMR$+$ (advantage: ${+}0.01687$). Task~2 exhibits similar trends with SDUM demonstrating $1.685 \pm 1.231$~dB (PSNR) and $0.02907 \pm 0.02333$ (SSIM) improvements, surpassing PromptMR+ by $1.060$~dB and $0.01827$, respectively. The head-to-head comparison reveals that SDUM outperforms PromptMR+ in $90.3\%$ of Task~1 cases and $93.9\%$ of Task~2 cases.

%% file: tables/training_infra.tex
\begin{table}[h]
\centering
\caption{\textbf{Scaling of SDUM with respect to unrolled depth $T$.} 
PSNR increases consistently with cascade count. Training infrastructure: NVIDIA H100 nodes (8 GPUs each).}
\label{tab:scaling_cascades}
\footnotesize
\setlength{\tabcolsep}{4pt}
\renewcommand{\arraystretch}{1.05}
\begin{tabular}{cccc}
\hline
$T$ & PSNR (dB) & Parameters & Training Infra \\
\hline
1  & 28.73 & 42.16\,M  & 1-node (8$\times$H100) \\
3  & 30.21 & 126.49\,M & 2-node (16$\times$H100) \\
6  & 32.09 & 252.97\,M & 4-node (32$\times$H100) \\
10 & 32.54 & 421.60\,M & 8-node (64$\times$H100) \\
18 & 33.18 & 758.91\,M & 8-node (64$\times$H100) \\
\hline
\end{tabular}
\end{table}

%% file: tables/inference_details.tex
\begin{table}[h]
\centering
\caption{\textbf{Inference computation details across different input sizes.} 
Reported values are averaged per slice on a single NVIDIA H100 GPU ($T{=}18$).}
\label{tab:inference_compute}
\footnotesize
\setlength{\tabcolsep}{4pt}
\renewcommand{\arraystretch}{1.05}
\begin{tabular}{lcc}
\hline
Input Size & Time (s) & Memory (GB) \\ \hline
$128 \times 128$ & 0.32 & 4.78 \\
$256 \times 256$ & 1.03 & 6.07 \\
$256 \times 512$ & 2.06 & 7.98 \\
$328 \times 512$ & 2.67 & 9.26 \\
$328 \times 640$ & 3.30 & 9.62 \\
$328 \times 768$ & 3.97 & 10.83 \\ \hline
\end{tabular}
\end{table}